\pdfoutput=1

\documentclass[11pt]{article}

\usepackage[]{acl}

\usepackage{times}
\usepackage{latexsym}

\usepackage[T1]{fontenc}

\usepackage[utf8]{inputenc}

\usepackage{microtype}
\usepackage{lscape} 
\usepackage{amssymb}
\usepackage{booktabs}
\usepackage{multirow}
\usepackage{nicematrix}
\usepackage{flushend}
\usepackage[export]{adjustbox}
\usepackage{enumitem}
\usepackage{tablefootnote}
\usepackage{xcolor}
\usepackage{booktabs}
\usepackage{pdflscape}
\usepackage{tabularx}
\usepackage{float}
\usepackage{placeins}
\usepackage{titlesec} 
\usepackage{hyperref}   
\usepackage[capitalise,nameinlink]{cleveref} 

\def\paperDraft{}

\ifdefined\paperDraft
 \def\JCGcomment#1{{\color{teal}[Javier: \textit{#1}]}}
 \def\AEcomment#1{{\color{purple}[Arash: \textit{#1}]}}
 \def\AScomment#1{{\color{green}[Alessandro: \textit{#1}]}}
 \def\HHcomment#1{{\color{orange}[Helen: \textit{#1}]}}

 \def\DraftOnly#1{{\color{red}[For Reference Only, do not read: \textit{#1}]}}

\else
 \def\JCGcomment#1{}
 
 \def\AEcomment#1{}
 
  \def\AScomment#1{}
 
 \def\HHcomment#1{}
 
 \def\DraftOnly#1{}
\fi

\title{\textit{`What are you referring to?'} Evaluating the Ability of Multi-Modal Dialogue Models to Process Clarificational Exchanges}

\author{Javier Chiyah-Garcia*~~Alessandro Suglia*$\dagger$~~Arash Eshghi*$\dagger$~~Helen Hastie*\\
  *Heriot-Watt University, Edinburgh, United Kingdom \\
  $\dagger$AlanaAI, Edinburgh, United Kingdom\\
  \texttt{\{fjc3,a.suglia,a.eshghi,h.hastie\}@hw.ac.uk} \\
}

\begin{document}
\maketitle
\begin{abstract}

\textit{Referential ambiguities} arise in dialogue when a referring expression does not uniquely identify the intended referent for the addressee. Addressees usually detect such ambiguities immediately and work with the speaker to \textit{repair} it using meta-communicative, Clarificational Exchanges (CE\footnote{Not to be confused with, but related to Clarification Ellipsis as used in e.g. \newcite{Fernandez.Ginzburg02a}}): a \textit{Clarification Request} (CR) 
and a response. 
Here, we argue that the ability to generate and respond to CRs imposes specific constraints on the architecture and objective functions of multi-modal, visually grounded dialogue models. We use the SIMMC 2.0 dataset to evaluate the ability of different state-of-the-art model architectures to process CEs, with a metric that probes the contextual updates that arise from them in the model. 
We find that language-based models are able to encode simple multi-modal semantic information and process some CEs, excelling with those related to the dialogue history, whilst multi-modal models can use additional learning objectives to obtain disentangled object representations, which become crucial to handle complex referential ambiguities across modalities overall\footnote{The source code and evaluation experiments are available at \href{https://github.com/JChiyah/what-are-you-referring-to}{https://github.com/JChiyah/what-are-you-referring-to}}.

\end{abstract}

\section{Introduction}

\begin{figure}[t]
\centering
\includegraphics[width=0.9\linewidth]{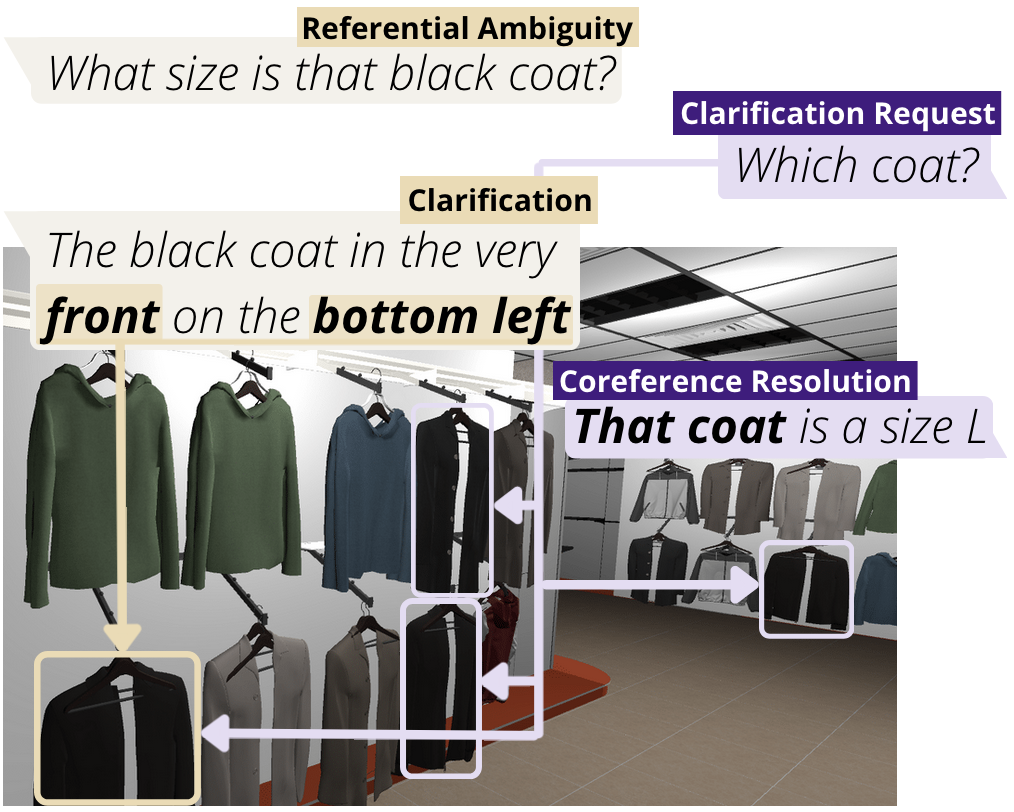}
\caption{Example referential ambiguity and clarification in SIMMC 2.0 dialogues.}
\label{fig:simmc2_coref_example}
\end{figure}

In dialogue, people work together on a moment by moment basis to achieve shared understanding and coordination \cite{Clark96,Clark.Brennan91,Goodwin81,Healey.etal18,Mills07}. A key mechanism people use to repair misunderstandings when they occur is via meta-communicative, clarificational exchanges (CE): a clarification request (CR) followed by a response (see ~\cref{fig:simmc2_coref_example}). CRs are a highly complex phenomenon: they are multi-modal \cite{Benotti.Blackburn21}, highly context-dependent with different forms and interpretations \cite{Purver04, Purver.Ginzburg04}, and can occur at different levels of communication on \citeauthor{Clark96}'s (\citeyear{Clark96}) joint action ladder \cite{Schlangen04,Benotti.Blackburn21}. But while the crucial role of generating and responding to CRs in dialogue systems has long been recognised \cite{San-Segundo.etal01,Rieser.Moore05,Rodriguez.Schlangen04,Rieser.Lemon06}, CRs still remain an understudied phenomenon \cite{Benotti.Blackburn21}, especially in the context of recent successes in multi-modal dialogue modelling \cite{suglia2021empirical, wang-etal-2020-vd, chen2020uniter, guo2022gravl, visdial2017, chen-etal-2021-multimodal, agarwal-etal-2020-history}. There is recent work related to identifying when to pose a CR \cite{Madureira.Schlangen23, zhu-2021, shi-etal-2022-learning}, but few evaluate the ability of models to process their responses \cite{10.1145/3462244.3479925, aliannejadi-etal-2021-building}.

In this paper, we use CRs as a testbed for studying and evaluating different neural dialogue model architectures (see also \newcite{Madureira.Schlangen23}). We focus on \textit{referential CRs} occurring at level three of \citeauthor{Clark96}'s (\citeyear{Clark96}) action ladder: that of \textit{understanding}. We provide a framework for evaluating how well multi-modal dialogue models are able to exploit referential CEs to resolve ambiguous referential descriptions. 
We use this framework to probe several state-of-the-art models proposed for the SIMMC 2.0 Challenge \cite{kottur-etal-2021-simmc} trained to resolve situated multi-modal coreferences with CEs found in the SIMMC 2.0 dataset itself. 

The results indicate that the ability of a model to exploit CRs to resolve referential ambiguities depends on the level of granularity of the model's cross-modal representations, i.e. how well information about different object attributes is represented. 
In particular, we find that the model that includes a training objective designed for predicting object attributes in a multi-task setup performs significantly better than the rest which was not optimised with this objective. This is in line with findings in \citet{suglia-etal-2020-compguesswhat} who show that having disentangled object representations \cite{bengio2013representation} allows models to better partition the search space of potential referents; and thereby better exploit effective object attributes in disambiguation.


\section{Dataset}

We used the SIMMC 2.0 dataset \cite{kottur-etal-2021-simmc}, which is a collection of multi-modal task-oriented dialogues, where both the system and the agent are situated in the same virtual environment. 
The dataset dialogues 
have a high degree of ambiguity 
and use rich referring expressions due to the overlap of many similar-looking objects (e.g., 5 red t-shirts in view); dialogues with references to multiple and previously discussed objects (mean 4.5 unique objects referenced per dialogue, SD: 2.4); and changing points of view throughout dialogues with partially observed objects. Thus, referential ambiguities in both the visual and conversational contexts are common. 
Furthermore, other common datasets do not contain coordination phenomena exhibited in SIMMC 2.0 (i.e. GuessWhat?! \cite{Vries_2017_CVPR}) or have a mixture of CRs which focuses solely on multi-modal referential ambiguities (e.g., Photobook \cite{haber-etal-2019-photobook}). 



\subsection{Dataset Details}\label{sec:dataset}

In the SIMMC 2.0 dataset \cite{kottur-etal-2021-simmc}, the agent acts as the shopping assistant to a user in a virtual shop. It encompasses the domains of fashion and furniture over 11,244 dialogues and it was collected using a mix of dialogue self-play and crowd-sourcing. The dataset is originally split into \texttt{train/dev/devtest/test-std} with 65\% / 5\% / 15\% / 15\% of the dialogues respectively. 

Each dialogue is complemented by images of the environment scene and rich metadata annotations. Some dialogues have multiple scene images with partially overlapping sets of objects, requiring models to carry over information from previous viewpoints. 
On average, dialogues have 5.2 utterance pairs (\textit{user-assistant}) and associated scenes have a mean of 27.6 objects, with some of them reaching up to a maximum of 141 items. \cref{tab:dialogue_sample1} shows a dialogue from the dataset, refer to \cref{ap:corpus_samples} for further samples.

\def\BeforeCR{\textbf{Before-CR\ }}
\def\AfterCR{\textbf{After-CR\ }}

\begin{table}[ht!]
\addtolength{\tabcolsep}{-0.4em}
\centering
\small
\begin{tabularx}{\linewidth}{r r X}

\toprule


& USR & \textit{Hello, do you have any jackets for me to look at?} \\
& SYS & \textit{Sure, what do you think of the light grey jacket hanging up high at the back left?} \\
& USR & \textit{Do you have anything with a similar size range to the black sweater beside the light grey jacket?} \\
& SYS & \textit{Sorry, I don't have anything similar to that} \\
\small{\BeforeCR}   & USR & \textit{What size is that sweater anyways?} [Referential Ambiguity] \\
\small{\textbf{CR\ }} & SYS & \textit{The \underline{black} one?} [Clarification Request]           \\
\small{\AfterCR}    & USR & \textit{Yes exactly} [Clarification]             \\
&  & \tiny{\texttt{-- Tags in CE: Individual Property}}                                            \\

\multicolumn{3}{c}{\includegraphics[width=0.65\linewidth]{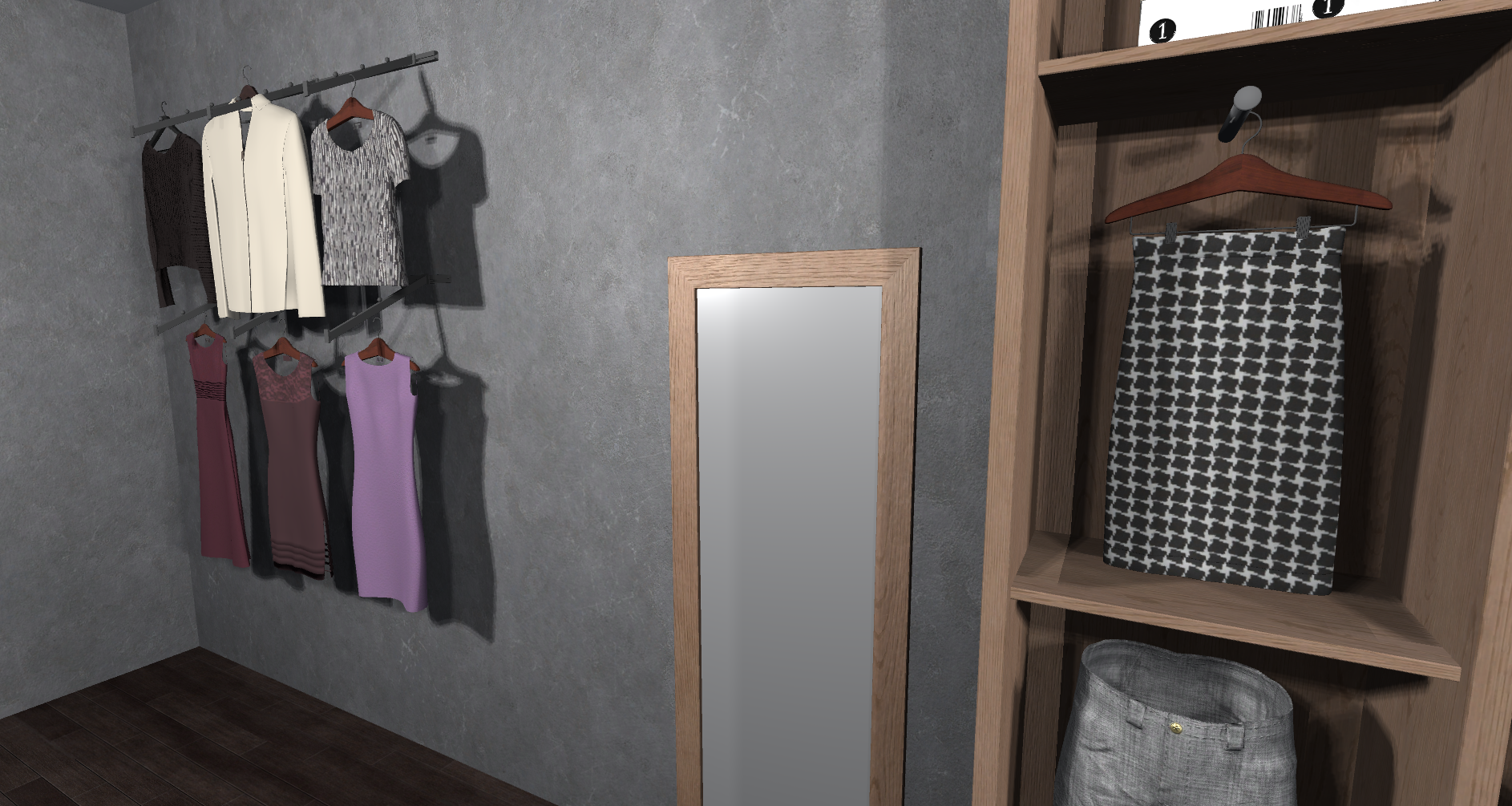}} \\

\bottomrule
\end{tabularx}
\caption{Sample dialogue with a CE from the SIMMC 2.0 dataset.}
\label{tab:dialogue_sample1}
\end{table}

Since the gold data from the \texttt{test-std} split is not available, we used the \texttt{devtest} data for our evaluation. Thus, some of the model object F1 scores may differ from their respective papers by a few decimals.

\subsection{CRs in SIMMC 2.0}

We focus on the clarificational sub-dialogues from the SIMMC 2.0 dataset. During the challenge, the dataset authors proposed several tasks, two of which are relevant here: Multi-modal Disambiguation (detecting whether the system has enough information to identify a unique object or is ambiguous) and Multi-modal Coreference Resolution (find the objects mentioned by the user). 
The dataset provides annotations that mark whether a turn is ambiguous or not, and which objects are referred to. Models were implicitly required to handle them as part of longer conversations, although the challenge did not explore clarifications in-depth.
We choose this dataset for studying CRs for two main reasons: 1) it contains complex multi-modal dialogues with gold labels for referential ambiguity; 2) it focuses on tasks such as disambiguation and coreference resolution in multi-modal settings that are directly related with the problem of CR resolution.

\subsection{Clarification Taxonomy}\label{sec:taxonomy}

To evaluate how models handle CEs, we need to understand their ability to exploit fine-grained contextual information across modalities beyond level three of \citeauthor{Clark96}'s (\citeyear{Clark96}) action ladder.
Therefore, we derive a taxonomy of different types of clarifications depending on the information or \textit{Disambiguating Property} exploited to resolve them: 1) \textbf{Individual Property}, such as object colour or state (i.e., \textit{``The \underline{red} jacket \underline{hanging}}''); 2) \textbf{Dialogue History}, such as referring to previously mentioned objects (i.e., \textit{``the one you \underline{recommended}''}); and 3) \textbf{Relational}, such as position or their relation to other objects in the scene (i.e., \textit{``the \underline{left} shirt, \underline{next to} the central rack''}).

These types are not mutually exclusive, and thus we often find that CRs are resolved with complementary information (i.e., \textit{``The \underline{green} dress on the \underline{right}''}). 
Refer to \cref{ap:corpus_samples} for discourse and taxonomy samples.

\section{Experimental Setup}

\subsection{Clarification Extraction and Tagging}

This section gives a summary of how we extracted the clarifications from the SIMMC 2.0 dataset using the gold annotations and tagged them using our taxonomy from Section \ref{sec:taxonomy}.

When a turn is annotated as ambiguous, the system generates a CR (e.g., \textit{``which one do you mean?''}). We label as \BeforeCR the user utterances preceding a CR (the user gave ambiguous information); whereas we label as \AfterCR the following user utterances that resolve the ambiguity. We obtain a subset of CEs (10\% of all system turns are CRs) which we use for the analysis. 
Finally, we use a keyword-based method to tag the disambiguating properties exploited for clarifications (cf. \cref{ap:cr_tagging}).

\subsection{Metrics}

We follow the SIMMC 2.0 evaluation protocol and measure coreference resolution performance using \textbf{Object F1}, derived as the mean of recall and precision for the predicted objects at each turn, as defined in \cite{kottur-etal-2021-simmc}.


Along with object F1, we look at the difference in F1 between the turns before and after a clarification. Intuitively, a model that can process clarifications will improve after one, reflecting a higher F1 in the set of turns after a CR. Similarly, the turns before a CR may perform poorly, signalling confusion or uncertainty in general. We take this as the \textbf{Relative Delta $\Delta$} to compare it across models.

\def\Mbaseline{{\small$Baseline_{\scriptscriptstyle GPT-2}$\ }}
\def\Mlanguage{{\small$GroundedLan_{\scriptscriptstyle GPT-2}$\ }}
\def\Mvisionlan{{\small$VisLan_{\scriptscriptstyle LXMERT}$\ }}
\def\Mrelational{{\small$MultiTask_{\scriptscriptstyle BART}$\ }}       

\subsection{Models}

For our evaluation, we selected publicly available state-of-the-art 
models that took part in the SIMMC 2.0 challenge
\footnote{Not all models were public and some had missing code or weights.}. We give the relevant model details below, but please refer to original papers for additional architectural information. 

\paragraph{Language-based} We use two GPT-2-based \cite{radford2019language} models: the  Baseline (\Mbaseline) from \citet{kottur-etal-2021-simmc} (36.6\% Object F1 ↑); and an improved version from one of the challenge participant teams \cite{hemanthage2022dstc10}, \Mlanguage (67.8\% F1 ↑). Both models are similar and treat the task as a generation task, and are jointly trained with other goals in the challenge (coreference resolution, dialogue state tracking and response generation).

\paragraph{Vision-and-Language} We take LXMERT-based \cite{Tan.Bansal19} model \cite{chiyah-garcia2022dstc10} (\Mvisionlan, 68.6\% F1 ↑) that combines the images from the visual scenes and the dialogue to predict the coreferenced objects at each turn. It extracts object attributes from a Detectron2 model \cite{wu2019detectron2} to use as textual descriptions along with the visual features. For each object in the scene, it outputs a probability for the object being referenced in that turn and selects those above a threshold.
This model is only trained on coreference resolution. 

\begin{table*}[ht]
\centering
\addtolength{\tabcolsep}{-0.35em}
\scriptsize
\begin{tabular}{@{}l|llc|llc|llc|llc@{}}

\toprule
\multicolumn{1}{r|}{\textbf{Model}}                                                       & \multicolumn{3}{c}{\textbf{\Mbaseline}}                    & \multicolumn{3}{c}{\textbf{\Mlanguage}}                    & \multicolumn{3}{c}{\textbf{\Mvisionlan}}                   & \multicolumn{3}{c}{\textbf{\Mrelational}}                  \\
\midrule

\textbf{Split}                                                       & \textbf{\BeforeCR} & \textbf{\AfterCR} & \textbf{$\Delta$} & \textbf{\BeforeCR} & \textbf{\AfterCR} & \textbf{$\Delta$} & \textbf{\BeforeCR} & \textbf{\AfterCR} & \textbf{$\Delta$} & \textbf{\BeforeCR} & \textbf{\AfterCR} & \textbf{$\Delta$} \\
\midrule

\textbf{All Turns}            & \multicolumn{2}{c}{34.3 (.01)}                     &                         & \multicolumn{2}{c}{67.8 (.01)}               &                   & \multicolumn{2}{c}{68.6 (.01)}           &                   & \multicolumn{2}{c}{74.0 (.01)}             &                   \\
\textbf{CR Turns}             & 36.4 (.01)               & 29.1 (.01)              & -20.1\%                 & 64.8 (.01)               & 67.7 (.01)        & +4.4\%            & 65.7 (.01)         & 69.2 (.01)          & +5.4\%            & 66.9 (.01)         & 74.3 (.01)          & +11.1\%           \\

\midrule

\multicolumn{2}{l}{\textbf{\tiny{Disambiguating Property}}} \\

\textbf{Individual Property}     & 35.4 (.02)         & 27.4 (.01)        & -22.7\%           & 65.0 (.02)         & 68.0 (.02)        & +4.6\%            & 65.1 (.02)         & 69.3 (.01)        & +6.4\%            & 68.0 (.02)         & \textbf{75.7} (.01)        & +11.3\%           \\
\textbf{Dialogue History} & 47.6 (.04)         & 43.7 (.04)        & -8.2\%            & 81.7 (.03)         & 82.1 (.03)        & +0.4\%            & 81.7 (.03)         & \textbf{84.6} (.03)        & +3.5\%            & 67.2 (.04)         & 75.7 (.04)        & +12.6\%           \\
\textbf{Relational Context}         & 32.9 (.02)         & 25.0 (.02)        & -24.1\%           & 62.4 (.02)         & 63.7 (.02)        & +2.1\%            & 62.7 (.02)         & 65.0 (.02)        & +3.7\%            & 66.5 (.02)         & \textbf{72.6} (.02)        & +9.1\%           \\

\bottomrule

\end{tabular}
\caption{Evaluation results for models at handling CEs with different disambiguating properties. Measured in \textbf{Object F1} ↑ (SD) and \textbf{Relative Delta $\Delta$}.}
\label{tab:results}

\end{table*}

\paragraph{Language-Vision-and-Relational} We use the model of the coreference challenge winner team \cite{lee-etal-2022-learning} (\Mrelational, 74\% F1 ↑), a BART-based model \cite{lewis-etal-2020-bart} trained to handle all challenge tasks. 
A pretrained ResNet model \cite{he2016deep} encodes each object along with its non-visual attributes, a learnable embedding that is later mapped to match the dimension of BART. 
The model is jointly optimised on multiple tasks, including several secondary tasks that enable learning disentangled object representations \cite{bengio2013representation} through object attribute slot prediction for each coreferenced object.
The object location is also encoded through the bounding box information and a location embedding layer. Finally, the canonical object IDs are used to ground relations between the object locations, the visual and non-visual attributes.

\section{Experiments}


\paragraph{Referential Ambiguities}
Firstly, we explore whether referential ambiguities are an issue for models and if clarifications are thus needed. 
From the initial two rows of \cref{tab:results}, we observe that, aside from the \Mbaseline model, all other models perform worse in turns \BeforeCR than when evaluating \textbf{All Turns}. This implies that indeed those utterances lack information to uniquely identify the referent objects, causing referential ambiguities for models and a lower object F1.

We also find that the F1 is higher in turns \AfterCR compared to turns \BeforeCR in all models but \Mbaseline. This suggests that models can at least process 
clarifications in some cases. The \Mvisionlan and \Mrelational models even benefit with increased performance in \AfterCR turns compared to \textbf{All Turns}.

Regarding the surprisingly high scores for the \Mbaseline in turns \BeforeCR and low for \AfterCR, we suspect that it is due to the model exploiting linguistic phenomena along with smart use of previously mentioned objects and their canonical IDs, as explained in \cite{chiyah-garcia2022dstc10}. The model's performance drops dramatically when it is crucial to carry over cross-turn information and ground it in dialogue which is required \AfterCR.

\paragraph{Disambiguating Properties}
Using the CR taxonomy (cf. \cref{sec:taxonomy}), we probe how models perform at exploiting different information with subsets of clarifications (bottom of \cref{tab:results}).

All models but the baseline show a similar performance in \BeforeCR turns that exploit an Individual Property. \Mlanguage and \Mvisionlan show a moderate F1 increase in the following \AfterCR turns, whereas \Mrelational obtains a more substantial improvement (+11.3\% $\Delta$). Individual object properties in this dataset relate to concepts in the visual context which may be difficult to see or complex to understand beyond colour or shape (e.g., long sleeve or folded). 

The \Mlanguage model implicitly encodes object attributes using a global object ID, which allows the model to learn latent information during training that carries over to evaluation sets (i.e. \texttt{<OBJ\_256>}). On the other hand, the \Mvisionlan model encodes colours and shapes explicitly using textual descriptions (i.e. \texttt{blue hoodie}) and implicitly in the visual region of interest features, which explains the slightly higher performance in these particular clarifications. 
However, the vision module of \Mvisionlan is not explicitly trained to detect complex properties, only attributes such as colours or shapes (i.e. blue hoodie), and is instead left to the visual features to represent this information.




The multi-task learning objectives of \Mrelational help the model obtain more fine-grained disentangled representations than using vision alone 
which helps in resolving ambiguities related to individual properties. \citet{suglia-etal-2020-compguesswhat} suggests that exploiting explicit object attributes reduces the potential referents and thus may also lead to improvements in solving CRs. 

\Mlanguage and \Mvisionlan models perform well when the clarifications are related to the dialogue context. Their initial F1 (+81\%) suggests that they are able to carry information across turns particularly well and may not even need a CR in these cases. Both models also improve in \AfterCR turns, with \Mvisionlan reaching the highest score for this category. 
On the other hand, \Mrelational improves its performance to 75.7\% F1 (+12.6\% $\Delta$), but it does not display the same ability to exploit the linguistic context as the other models.
This is likely due to the multi-task formulation involving specific loss functions which focus on visual and relational information only. Thus, the model obtains strong visual and relational object representations, whilst affecting the quality of BART's pre-trained language representations.




Relational clarifications seem to be the most difficult type to process for models, with the lowest F1 scores overall.
The \Mrelational model is able to exploit this information considerably better than the other models and improves by a +9.1\% to 72.6\%. This is an important strength of the model which extends its ability to encode visual attributes of the objects with information about the relationships between the objects in the scene. For instance, this model is able to capture the positions of the objects in the scene and how they relate to each other. The \Mvisionlan model encodes positional information such as bounding box coordinates too, but it is not able to learn from them \cite{chiyah-garcia2022dstc10}. This is justified by previous research by \cite{Salin_Farah_Ayache_Favre_2022} that shows how multi-modal models struggle with concepts such as position, and that they rely on language bias instead. 


\section{Conclusion}

Referential ambiguities are common in situated human conversations. We sometimes cannot fully understand or identify a referred object or event, and thus we engage in clarification exchanges to resolve the ambiguity. 
In this paper, we analyse how several state-of-the-art models treat clarifications in situated multi-modal dialogues using the SIMMC 2.0 dataset. We classify the types of clarifications by the disambiguating property exploited 
and then evaluate the models with subsets of the data. 

We find that language-based models perform well, yet struggle to benefit from clarifications. On the other hand, vision seems to be an important (but not essential) addition for models, which helps processing multi-modal CEs.
Paired with a strong dialogue context, these types of models can perform reasonably well and carry information across turns to better handle clarifications.
Finally, encoding relations between objects and their locations, and using additional learning objectives to predict attribute slots seems the strongest architecture for models to handle CEs. 

Based on these results, to create improved models that can resolve referential ambiguities in situated dialogues, we need \textit{holistic object-centric representations} that contain information about attributes and properties~\citep{seitzer2022bridging}, and that can \textit{dynamically} change to reflect the information exchanges available in the dialogue context.

\section*{Acknowledgements}

Chiyah-Garcia’s PhD is funded under the EPSRC iCase  with Siemens (EP/T517471/1). This work was also supported by the EPSRC CDT in Robotics and Autonomous Systems (EP/L016834/1).  

\bibliography{anthology,custom,refs,all}
\bibliographystyle{acl_natbib}

\clearpage
\newpage
\newpage

\appendix






\section{Additional CR Details}\label{ap:cr_tagging}

\subsection{Clarification Tagging Method}

The algorithm for CR tagging is based on manual annotations using the dev set, and then creating a set of keywords and regexes that would automatically find the disambiguating property used. \textbf{Individual Properties} include mentions of: colour (\textit{blue}), object types (\textit{jacket}), style (\textit{floral}), brand names (\textit{Yogi Fit}), states (\textit{folded}) and other (\textit{long-sleeve}). The metadata provides all of this information that we use as keywords.
Other information such as \textbf{Relational Context} was based on positional keywords (\textit{left, top}) and relational with objects (\textit{next to}) or the scene (\textit{farthest}). \textbf{Dialogue History} was based on linguistic cues and the use of common structures (i.e. \textit{``...in my cart''}, \textit{``you mentioned''}). 
We left some CRs as unclassified ($<0.06$\% of the data) because they do not provide any meaningful additional information or are out of scope
(i.e., \textit{``What is that lamp made of? - Sorry, which one? - I'm not sure, I think it's a lamp''}). 

During tagging, we applied the algorithm to each clarification, including both the system CR and the user response, as the information is sometimes scattered across turns (i.e., see SYS CR in Dialogue Sample 1).
We manually verified this method and designed unit tests to ensure correct classification as the tagging algorithm increased in complexity. 


\subsection{Ambiguity in SIMMC 2.0 Scenes}

\begin{table}[hb!]
\centering
\small
\begin{tabular}{l c c}
\toprule

               & \multicolumn{2}{c}{{\textbf{Mean Candidate Objects (SD)}}} \\
\textbf{Split} & \textbf{\texttt{Type}} & \textbf{\texttt{Colour}} \\
\midrule

 All  Turns & 3.10  (5.16) & 2.58  (4.23)  \\ 
 CR  Turns & 5.41  (5.62) & 4.53  (4.63)  \\ 

\multicolumn{2}{l}{\textbf{\scriptsize{\ \ Disambiguating Property}}} \\

 Individual  Property & 5.52  (5.69) & 4.57  (4.77)  \\ 
 Dialogue  History & 4.57  (4.82) & 4.95  (5.25)  \\ 
 Relational  Context & 5.78  (5.91) & 4.63  (4.66)  \\

\bottomrule
\end{tabular}
\caption{Statistics about the level of ambiguity in SIMMC 2.0 based on candidate objects of the same \textbf{Type} (\textit{jacket, t-shirt}...) or \textbf{Colour} (\textit{blue, yellow...}). These properties alone are not enough to uniquely identify an object in most cases.}
\end{table}

\vfill\break

\section{Corpus Examples}\label{ap:corpus_samples}
\vspace{-1em}

\begin{table}[h!]
\addtolength{\tabcolsep}{-0.4em}
\centering
\scriptsize
\begin{tabularx}{\linewidth}{r r X}

\toprule

\multicolumn{3}{c}{\small\textbf{Dialogue Sample 1}} \\

& USR & \textit{Hello, do you have any jackets for me to look at?} \\
& SYS & \textit{Sure, what do you think of the light grey jacket hanging up high at the back left?} \\
& USR & \textit{Do you have anything with a similar size range to the black sweater beside the light grey jacket?} \\
& SYS & \textit{Sorry, I don't have anything similar to that} \\
\tiny{\BeforeCR}   & USR & \textit{What size is that sweater anyways?} [Referential Ambiguity] \\
\tiny{\textbf{CR\ }} & SYS & \textit{The \underline{black} one?} [Clarification Request]           \\
\tiny{\AfterCR}    & USR & \textit{Yes exactly} [Clarification]             \\
&  & \tiny{\texttt{-- Tags in CE: Individual Property}}                                            \\

\multicolumn{3}{c}{\includegraphics[width=0.65\linewidth]{cloth_store_1416238_woman_3_11.png}} \\

\midrule
\multicolumn{3}{c}{\small\textbf{Dialogue Sample 2}} \\

& USR & \textit{I need some help picking out a jacket.} \\
& SYS & \textit{Do you like the look of that solid black one hanging in the bottom row up front on the right?} \\
& USR & \textit{I do like the look of that one. Will you please put that in my cart?} \\
& SYS & \textit{Sure! I'll add that black jacket to your cart right now.} \\
& USR & \textit{Now could we look for a grey coat?} \\
& SYS & \textit{There's a grey coat hanging directly above the black jacket you like.} \\

\tiny{\BeforeCR}     & USR & \textit{Actually first I'd like to know which brand makes that jacket. If you have any data on what other customers think of it, that'd be good to hear too.} [Referential Ambiguity] \\
\tiny{\textbf{CR\ }} & SYS & \textit{Sorry, which jacket specifically are you curious about?} [Clarification Request] \\
\tiny{\AfterCR}      & USR & \textit{The \underline{black} one \underline{up front next to} the skirt} [Clarification] \\
                     &     & \tiny{\texttt{-- Tags in CE: Individual Property, Relational}} \\

& SYS & \textit{That jacket is a Yogi Fit original, and other buyers have rated it a 4.0 on average} \\
& USR & \textit{Are there any super ritzy options with stellar reviews?} \\

& SYS & \textit{I don't see any, sorry. Is there something else I can help you with?} \\
& USR & \textit{I'd like to find a dress in extra large.} \\
& SYS & \textit{I have a feeling you'll like the dresses on this first rack up front and center. Would you wear the purple one on the left, the brown one on the right, or the grey one in front?} \\

\tiny{\BeforeCR}     & USR & \textit{Woops, I didn't even ask you what sizes that jacket came in.} [Referential Ambiguity] \\
\tiny{\textbf{CR\ }} & SYS & \textit{Which jacket do you want to know the sizes of?} [Clarification Request] \\
\tiny{\AfterCR}      & USR & \textit{That \underline{black} one that \underline{I had you put in my cart}} [Clarification] \\
                     &     & \tiny{\texttt{-- Tags in CE: Individual Property, Dialogue History}} \\

\multicolumn{3}{c}{\includegraphics[width=0.65\linewidth]{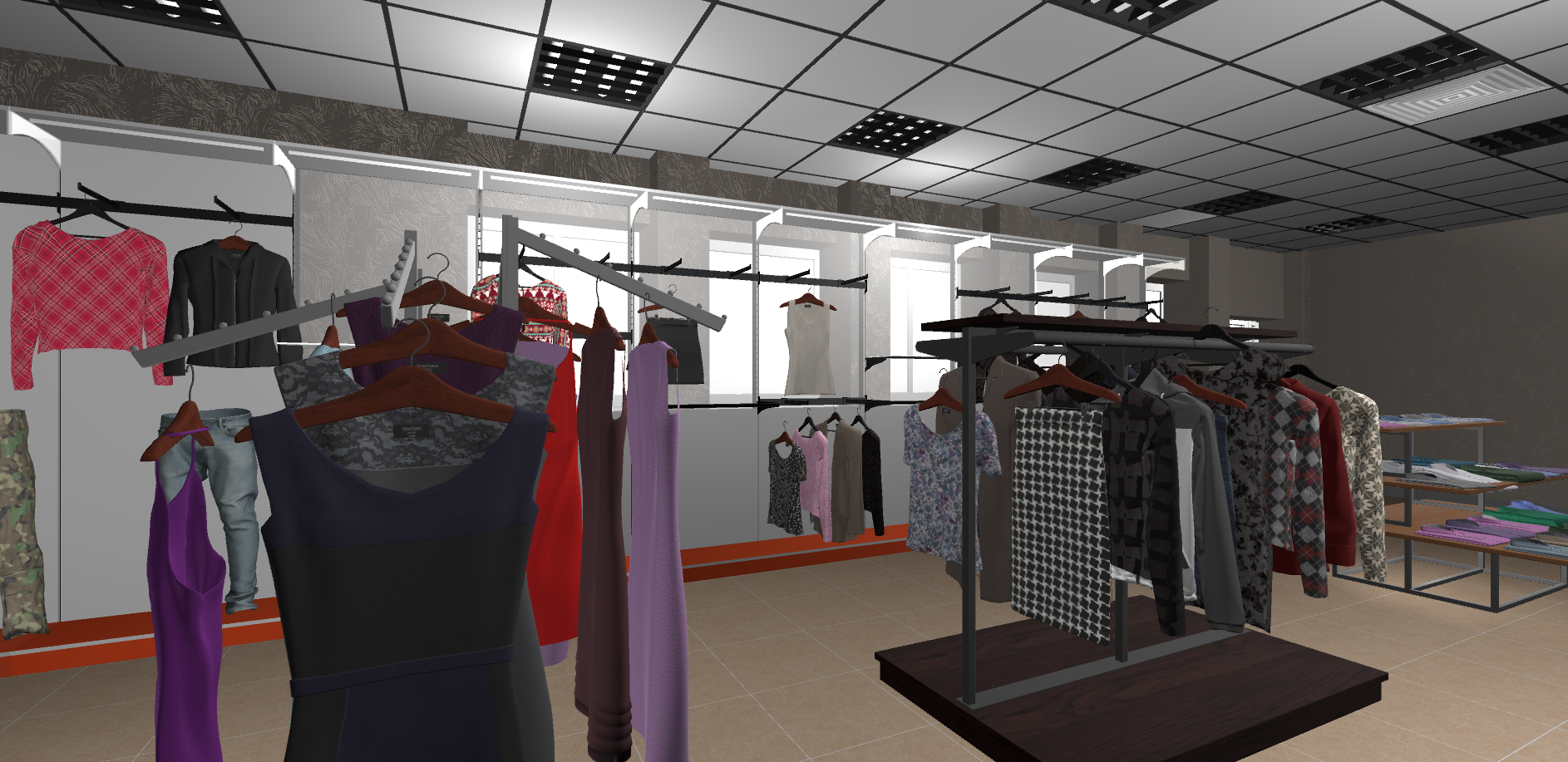}} \\
\midrule

\multicolumn{3}{c}{\small\textbf{Dialogue Sample 3}} \\

& USR & \textit{Can you find me a black blouse?} \\
& SYS & \textit{What about this black blouse in the shelf display?} \\
\tiny{\BeforeCR}   & USR & \textit{What's the size of the grey shirt and the grey and brown shirt?} [Referential Ambiguity] \\
\tiny{\textbf{CR\ }} & SYS & \textit{Which shirts are you referring to?} [Clarification Request]           \\
\tiny{\AfterCR}    & USR & \textit{I mean the \underline{grey} \underline{shirt} \underline{on the table} display and the \underline{grey} and \underline{brown} \underline{shirt} \underline{next to it}} [Clarification]             \\
&  & \tiny{\texttt{-- Tags in CE: Individual Property, Relational}}                                            \\

\multicolumn{3}{c}{\includegraphics[width=0.65\linewidth]{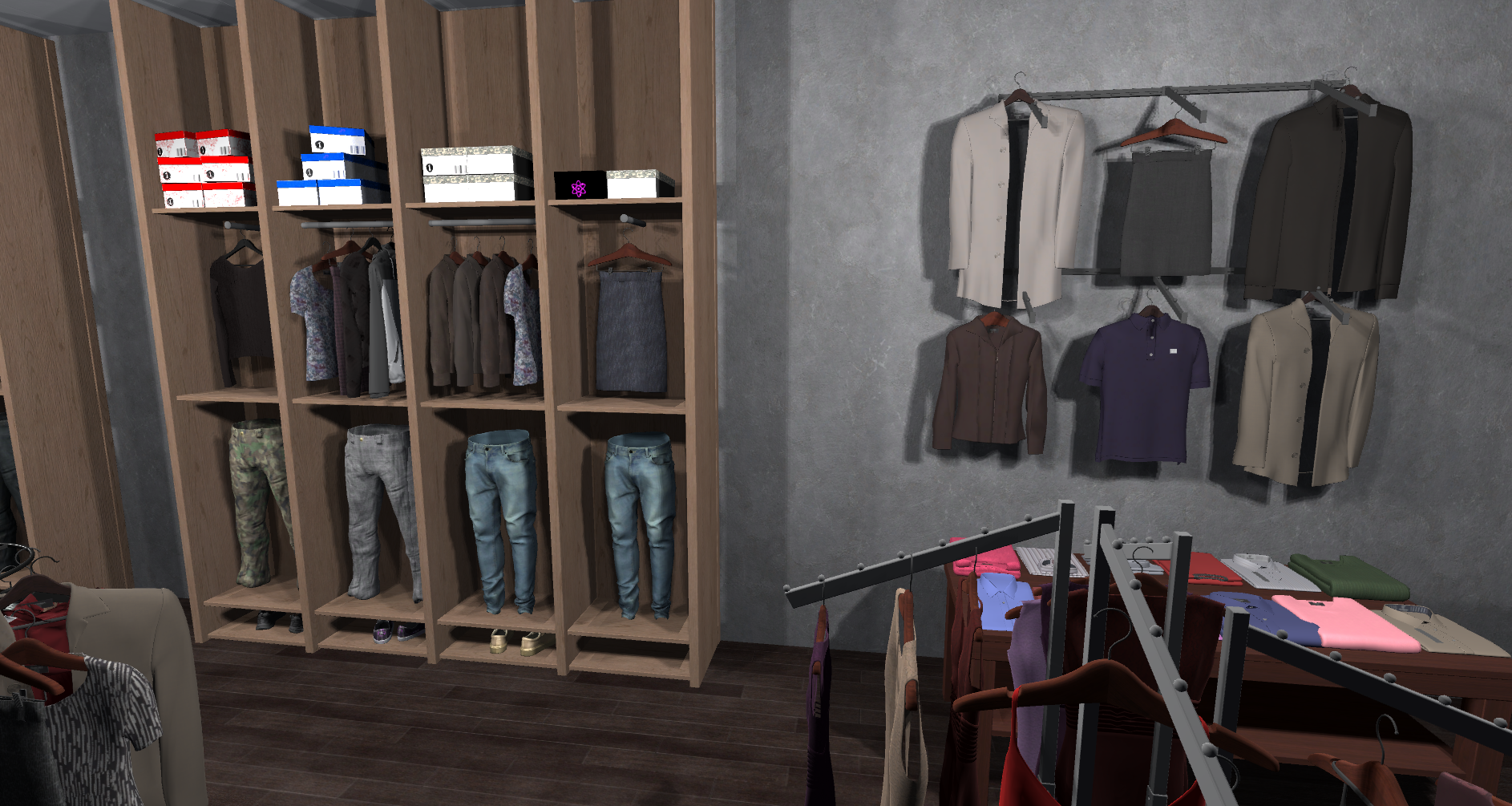}} \\

\bottomrule
\end{tabularx}
\end{table}

\vspace{-0.9em}

\clearpage





\end{document}